\documentclass{article}

 \usepackage[preprint]{neurips_2025}

\usepackage[utf8]{inputenc} 
\usepackage[T1]{fontenc}    
\usepackage{hyperref}       
\usepackage{url}            
\usepackage{booktabs}       
\usepackage{amsfonts}       
\usepackage{bbding}       %
\usepackage{nicefrac}       
\usepackage{microtype}      
\usepackage[dvipsnames]{xcolor}         
\usepackage{graphicx}
\usepackage{diagbox}
\usepackage{multirow}
\usepackage{multicol}
\usepackage{makecell}
\usepackage{amsmath}
\usepackage{float}
\usepackage[table]{xcolor}
\newcommand{\ours}{TableGPT-R1}
\usepackage{graphicx}
\usepackage{subcaption}

\title{
\raisebox{-0.25\height}{\includegraphics[width=1cm]{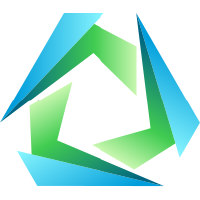}} \ours{}: Advancing Tabular Reasoning \\Through Reinforcement Learning
}

\author{%
  Saisai Yang\thanks{Joint first authors.} \quad Qingyi Huang\footnotemark[1] \quad Jing Yuan\footnotemark[1] \quad Liangyu Zha\footnotemark[1] \quad Kai Tang\footnotemark[1] \quad Yuhang Yang\footnotemark[1] \\
  \textbf{Ning Wang \quad Yucheng Wei \quad Liyao Li \quad Wentao Ye \quad Hao Chen} \\
  \textbf{Tao Zhang \quad Junlin Zhou \quad Haobo Wang\thanks{Joint corresponding authors.} \quad Gang Chen\footnotemark[2] \quad Junbo Zhao\footnotemark[2]}\\
  \\
  Zhejiang University \quad Institute of Computing Innovation, Zhejiang University \\
}

\begin{document}

\maketitle
\setcitestyle{numbers,square}

\begin{abstract}
\vspace{-5pt}
Tabular data serves as the backbone of modern data analysis and scientific research. While Large Language Models (LLMs) fine-tuned via Supervised Fine-Tuning (SFT) have significantly improved natural language interaction with such structured data, they often fall short in handling the complex, multi-step reasoning and robust code execution required for real-world table tasks. Reinforcement Learning (RL) offers a promising avenue to enhance these capabilities, yet its application in the tabular domain faces three critical hurdles: 
the scarcity of high-quality agentic trajectories with closed-loop code execution and environment feedback on diverse table structures, the extreme heterogeneity of feedback signals ranging from rigid SQL execution to open-ended data interpretation, and the risk of catastrophic forgetting of general knowledge during vertical specialization.
To overcome these challenges and unlock advanced reasoning on complex tables, we introduce \textbf{TableGPT-R1}, a specialized tabular model built on a systematic RL framework. Our approach integrates a comprehensive data engineering pipeline that synthesizes difficulty-stratified agentic trajectories for both supervised alignment and RL rollouts, a task-adaptive reward system that combines rule-based verification with a criteria-injected reward model and incorporates process-level step reward shaping with behavioral regularization, and a multi-stage training framework that progressively stabilizes reasoning before specializing in table-specific tasks. Extensive evaluations demonstrate that TableGPT-R1 achieves state-of-the-art performance on authoritative benchmarks, significantly outperforming baseline models while retaining robust general capabilities. Our model is available at \url{https://huggingface.co/tablegpt/TableGPT-R1}.
\vspace{-7pt}
\end{abstract}
\vspace{-10pt}
\begin{figure*}[htbp]
    \centering
    \begin{subfigure}{0.47\textwidth}
        \centering
        \includegraphics[width=\linewidth]{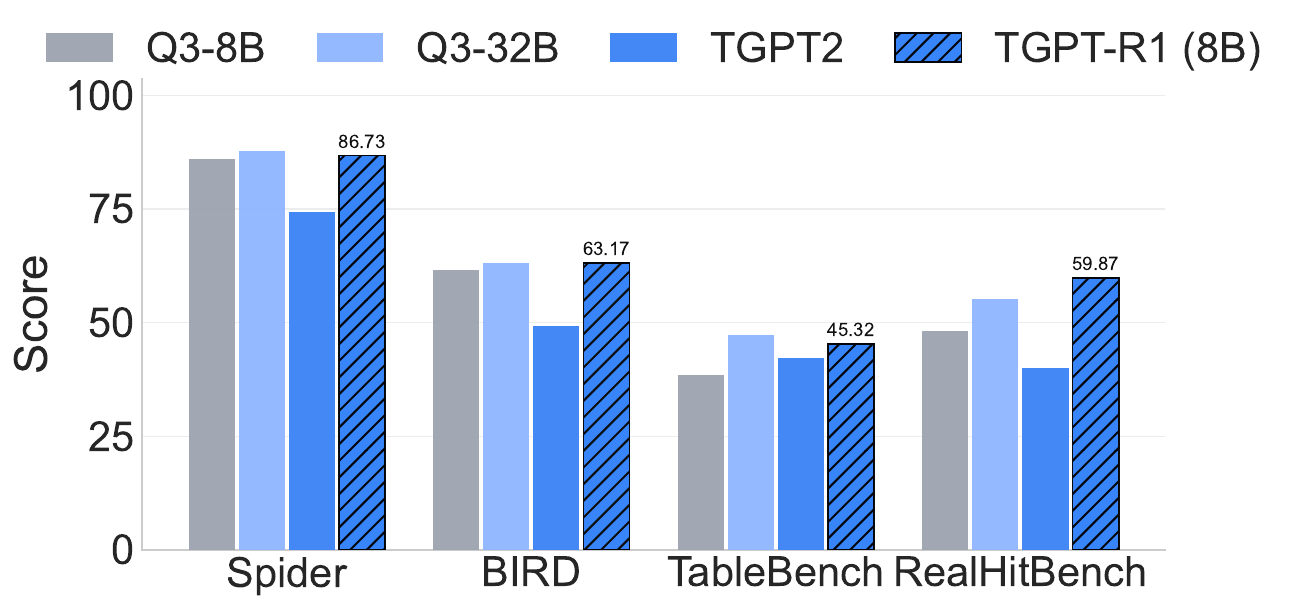}
        \caption{\textbf{Table benchmarks.}}
        \label{fig:intro_table_perf}
    \end{subfigure}
    \hfill
    \begin{subfigure}{0.47\textwidth}
        \centering
        \includegraphics[width=\linewidth]{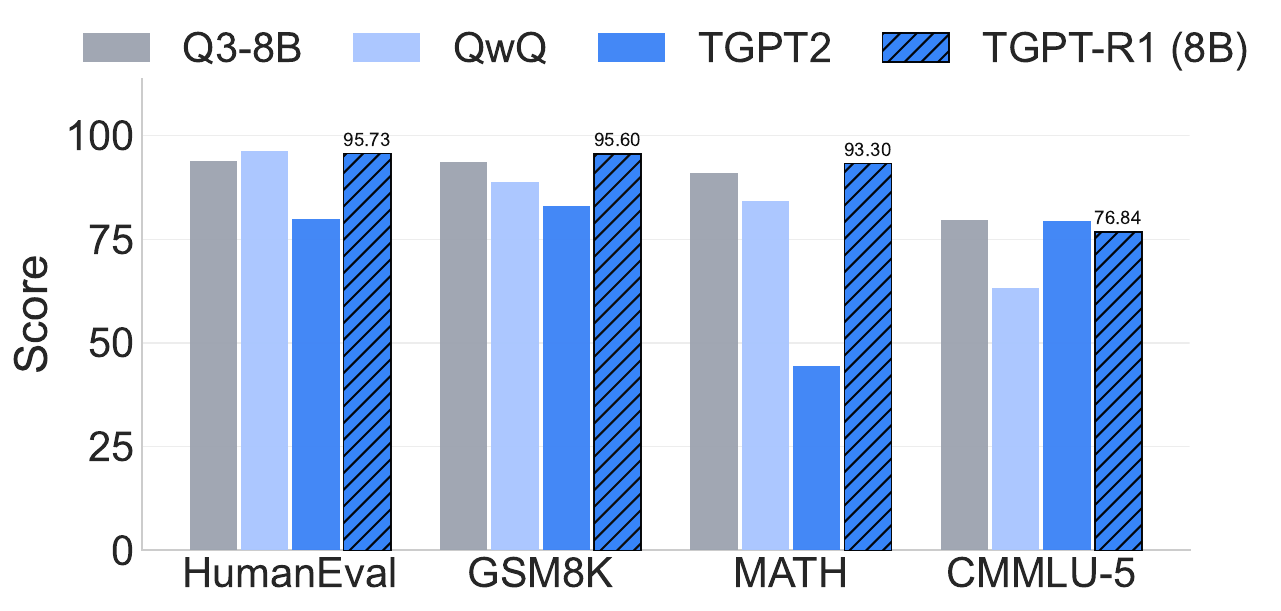}
        \caption{\textbf{General benchmarks.}}
        \label{fig:intro_general_perf}
    \end{subfigure}
    \caption{Performance comparison of TableGPT-R1 on tabular (left) and general (right) benchmarks. \textbf{Q3-8B}: Qwen3-8B; \textbf{Q3-32B}: Qwen3-32B; \textbf{TGPT2}: TableGPT2-7B; \textbf{TGPT-R1}: TableGPT-R1-8B.}
    \label{fig:intro_performance}
\end{figure*}
\clearpage
\tableofcontents
\clearpage

\section{Introduction}
Tabular data provide a foundational structure for storing, organizing, and presenting information~\cite{borisov2022deep,sahakyan2021explainable}, underpinning data analysis and data engineering, which are essential in fields like business intelligence and scientific research~\cite{add1}. The advent of Large Language Models (LLMs)~\cite{guo2025deepseek,bai2023qwen,comanici2025gemini,achiam2023gpt} has revolutionized how users interact with this structured data, shifting the paradigm from manual coding to intuitive natural language interaction~\cite{2024Large}. While Supervised Fine-Tuning (SFT) establishes basic instruction-following capabilities~\cite{wu2025llm,wei2021finetuned}, it often falls short in handling the complex, multi-step reasoning and robust code execution required for real-world table tasks~\cite{add2}. Reinforcement Learning (RL) offers a promising path to enhance these reasoning capabilities and generalization power~\cite{rafailov2023direct,kaufmann2024survey,dapo,add3}. However, applying RL to the tabular domain introduces three distinct challenges. First, \textbf{Data Scarcity for Table Agents} remains a significant hurdle. Unlike general chat, tabular tasks require the model to master a closed-loop of thinking, coding, and execution with environment feedback on tables of diverse forms. Traditional corpora rarely contain these precise, executable trajectories (Question-Code-Observation-Answer), and real-world tables possess diverse structures that general data cannot cover. Second, \textbf{Feedback Heterogeneity} complicates the reward mechanism. Tabular tasks exhibit a high degree of heterogeneity in feedback signals, spanning a spectrum from rigid code execution (where correctness is binary and strict) to open-ended data interpretation (where quality is subjective)~\cite{jin2022survey,hu2023chatdb}. Moreover, agentic trajectories induce long-horizon credit assignment, making pure terminal supervision brittle; process-level step reward shaping is often necessary to stabilize optimization. A single reward mode cannot effectively guide the model across these disparate objectives. Third, \textbf{Catastrophic Forgetting} poses a risk to model performance~\cite{van2024continual}. Balancing domain-specific expertise with general intelligence is difficult, as naive training strategies often lead to catastrophic forgetting of general reasoning abilities or instability during the intensive optimization of tabular skills.

To address these challenges, we introduce \textbf{TableGPT-R1}, a specialized tabular model built upon a systematic RL framework. Our approach is grounded in three core technical pillars corresponding to the challenges above.
To systematically tackle the issue of data scarcity, we develop a comprehensive \textbf{Data Engineering Pipeline} that encompasses data acquisition, synthetic agentic generation, and rigorous quality control. We first aggregate heterogeneous raw data from diverse sources---including general instruction corpora, table-specific benchmarks, and agent interaction logs. To enable robust code execution, we synthesize high-quality agentic trajectories that explicitly model the closed-loop process of reasoning, coding, and execution with environment feedback, and we leverage these trajectories for both supervised alignment and RL rollouts. To make the tabular setting concrete, we illustrate a typical agentic trajectory for table analysis. Given a user query and a table (often only partially visible in context), the model must plan, execute code to retrieve the necessary evidence, and then synthesize the final answer grounded in execution results. During training, we represent this process with a structured format that interleaves reasoning, code execution, and observations, enabling closed-loop learning for complex table tasks. Furthermore, we employ a difficulty-aware stratification strategy to balance sample complexity, ensuring the model receives a structured and learnable curriculum that covers both simple queries and complex analytical tasks.


To manage the heterogeneity of feedback signals, we design a \textbf{Task-Adaptive Reward System} that adaptively routes tasks to the most appropriate verification pathway. For open-ended reasoning tasks lacking definitive ground truth, we employ a Criteria-Injected Reward Model, which is trained via a rigorous pipeline involving teacher-student distillation and reinforcement learning to generate objective, criteria-based evaluations rather than subjective scalar scores. Conversely, for deterministic tasks with clear labels (e.g., SQL generation, math problems), we utilize a Rule-based Reward Function that integrates strict execution verification with composite regularization terms to enforce behavioral norms. In addition, we incorporate lightweight process-level step reward shaping during agentic RL rollouts to provide denser guidance for long-horizon optimization. This dual-track system ensures precise guidance across the full spectrum of tabular tasks, while robust constraints effectively mitigate reward hacking behaviors such as opportunistic plotting.

\begin{figure*}[!t]
    \centering
    \includegraphics[width=1.0\linewidth]{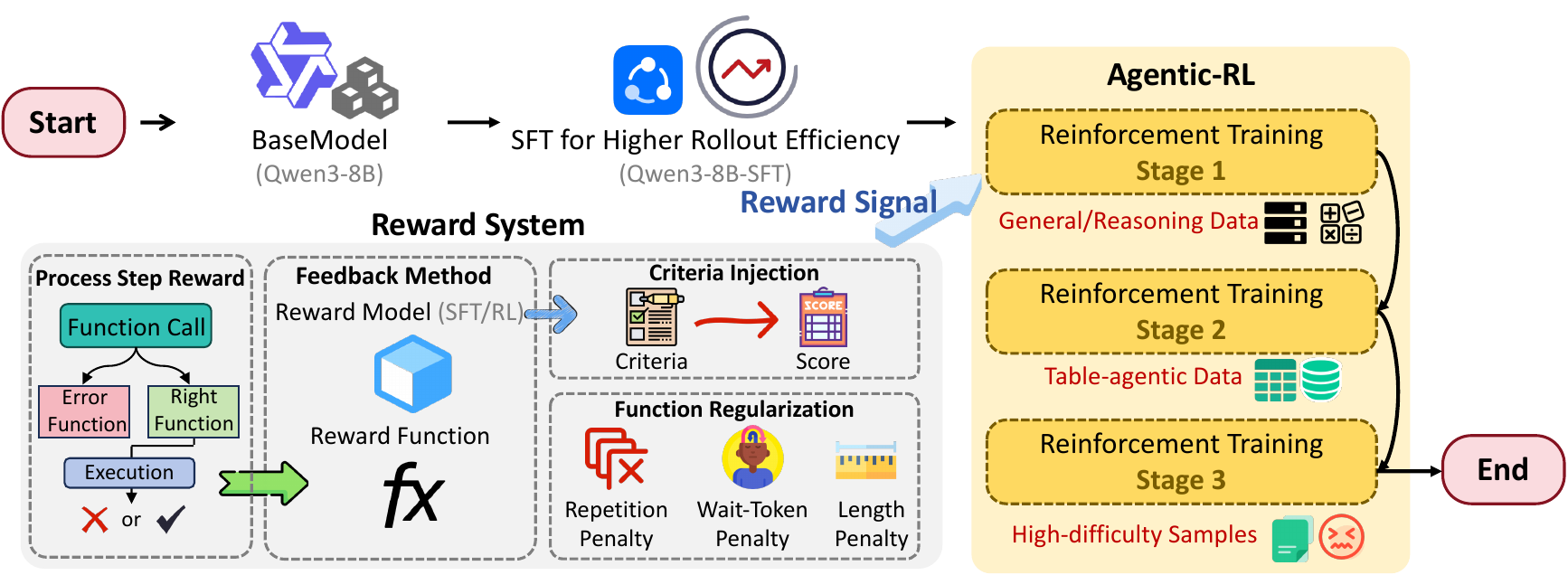}
    \caption{Overall framework of TableGPT-R1.}
    \label{fig:overall}
\end{figure*}

Finally, to resolve the optimization instability inherent in vertical domain adaptation, we implement a \textbf{Multi-Stage Training Strategy}. This curriculum-based approach begins with an SFT warm-up for format alignment, followed by a phased reinforcement learning process. We first stabilize general reasoning capabilities using broad-domain data, then progressively specialize in complex table manipulation, and ultimately focus on mining and conquering hard samples. This staged progression allows the model to incrementally build domain expertise without suffering from catastrophic forgetting of its foundational intelligence.

We conducted comprehensive evaluations across a wide spectrum of benchmarks, spanning both specialized tabular tasks and general reasoning domains, to validate both the superior tabular performance and the robust general capabilities of TableGPT-R1. In tabular domains, the model demonstrates exceptional capabilities in structure understanding, code-based analysis, and complex reasoning, achieving an average improvement of 11.32\% over TableGPT2-7B~\cite{su2024tablegpt2}. 
Simultaneously, extensive testing on general benchmarks confirms that TableGPT-R1 retains strong foundational intelligence. Despite slight regressions in specific tasks, the model exhibits an average performance gain of $1.01\%$ compared to the base model (Qwen3-8B~\cite{yang2025qwen3}), demonstrating that it successfully balances domain specialization with robust general capabilities.
To ensure a rigorous assessment, we benchmark against a diverse array of baselines, including leading open-source models and top-tier proprietary systems, proving that TableGPT-R1 delivers dominant domain expertise while maintaining the versatility expected of a general-purpose assistant.

\section{Training Data Construction}

Tabular data is widely used across domains as a structured format for storing, organizing, and presenting information. However, its row-column layout and operation-dependent semantics pose unique challenges for language models. Effective table reasoning demands process-level supervision—yet existing resources fall short: general datasets rarely involve tables, while table-specific benchmarks (e.g., TableQA or Text-to-SQL) typically provide only input–output pairs, omitting the intermediate reasoning and execution steps. To bridge this gap, we synthesize a high-quality agentic dataset featuring executable reasoning trajectories that explicitly model the full analytical workflow, serving as the foundation for supervised fine-tuning (SFT) and reinforcement learning (RL).

\subsection{Data Collection}

\begin{figure*}[!t]
    \centering
    \includegraphics[width=1.0\linewidth]{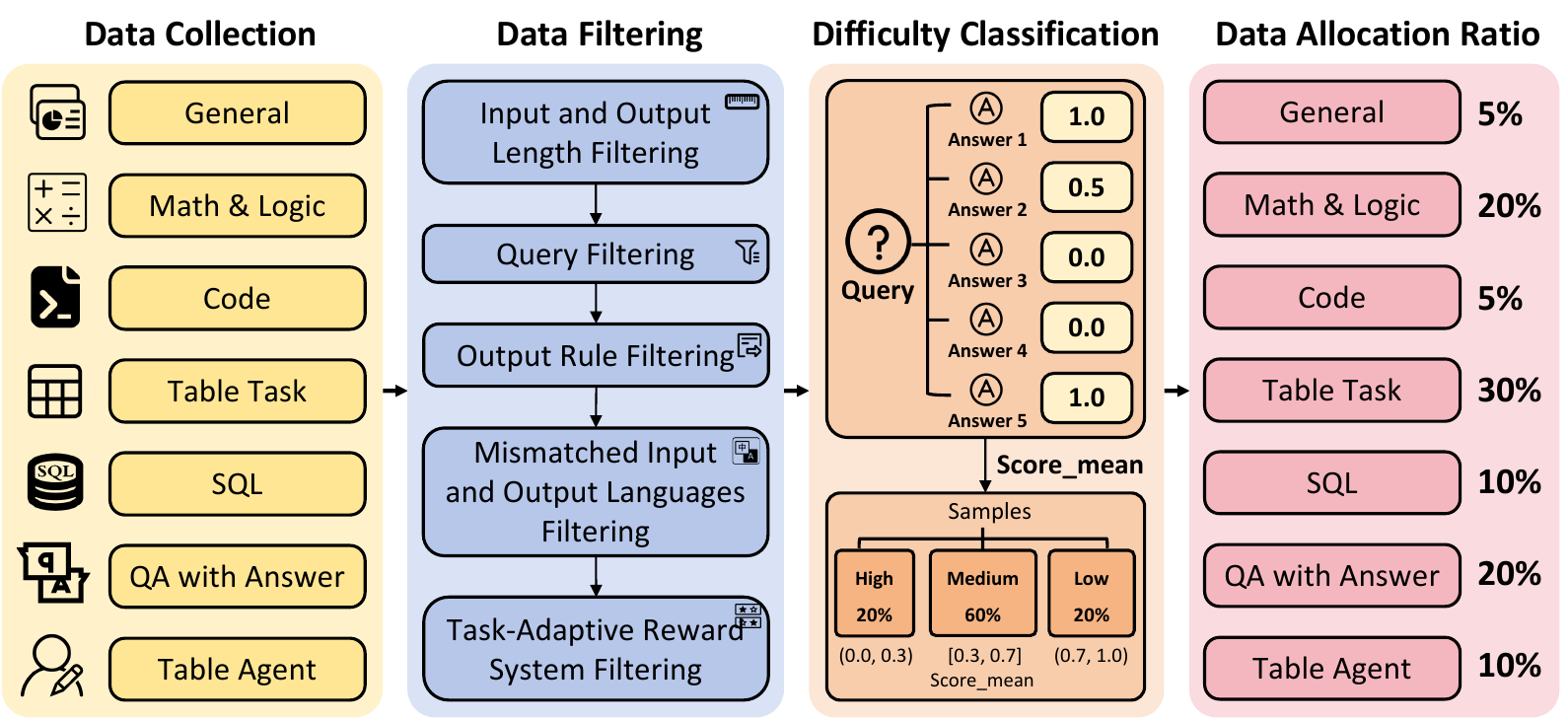}
    \caption{Data construction pipeline and final composition for TableGPT-R1.}
    \label{fig:agent_workflow}
\end{figure*}

We collect training data from a diverse set of public sources to ensure broad coverage of both tabular content and associated user queries or instructions. Specifically, we gather table-question pairs and table-related tasks from open repositories on GitHub, community benchmarks on Hugging Face and ModelScope, and web-crawled resources containing structured tables paired with natural language questions, SQL queries, or code snippets for data manipulation. In addition, to preserve and enhance the model’s general-purpose reasoning capabilities, we incorporate high-quality general-domain datasets spanning open-domain question answering, mathematical problem solving, and code generation. The collected data span a wide range of domains, including but not limited to government records, academia, manufacturing, finance, education, and healthcare, reflecting real-world usage across various scenarios.

\subsection{Data Filtering} 
After data collection, we apply a two-stage filtering pipeline to ensure high-quality training data, including input-based filtering and output-based filtering. 

\paragraph{Input-based Filtering} we first remove irrelevant, inconsistent or erroneous samples through: (1) \textit{Length Filtering}: length truncation ($\le$8,192 tokens) to fit the model’s context window and avoid verbosity; (2) \textit{Deduplication}: near-duplicate queries are detected and removed by MinHash across the entire corpus; (3) \textit{Rule-based Cleaning}: remove noisy or erroneous samples such as garbled text, harmful content and model identity leakage; and (4) \textit{Language Alignment}: discard input–output pairs with mismatched languages (e.g., a Chinese question paired with an English response). 

\paragraph{Output-based Filtering} we leverages a \textbf{Task-Adaptive Reward System} (detailed in Section~\ref{sec:Task-Adaptive Reward System}) to ensure the training data's answer quality. Specially, we first scores each sample on a 0–1 scale based on the quality of its reference response, retaining only those with scores $\ge$ 0.7, thereby preserving the reliability of the original data.

To enhance training efficacy, we further shape the difficulty distribution of the final training set. For each retained sample, we estimate its difficulty by generating five independent responses using the current model and scoring them with the reward system. Denoting the score of the $i$-th sampled response as $m^{(i)}$, the average score $\bar{m} = \frac{1}{5} \sum_{i=1}^{5} m^{(i)}$ determines the sample’s difficulty level. Based on $\bar{m}$, samples are categorized as:

\begin{itemize}
    \item \textbf{High difficulty}: $\bar{m} < 0.3$
    \item \textbf{Medium difficulty}: $0.3 \leq \bar{m} \leq 0.7$
    \item \textbf{Low difficulty}: $\bar{m} > 0.7$
\end{itemize}

We then resample the entire corpus to enforce a target distribution of \textbf{20\% high}, \textbf{60\% medium}, and \textbf{20\% low} difficulty samples.
This balanced difficulty distribution maximizes training efficiency by focusing on informative, learnable examples while maintaining sufficient coverage of both foundational and challenging cases, thereby promoting stable learning and robust generalization.

\subsection{Data Composition}
We iterated through multiple rounds of data composition and refinement, adjusting the mixture and characteristics of the training data in close alignment with our model training cycles. The final version of our dataset integrates three key components:
(1) \textbf{Agentic data (see Section~\ref{sec:agentic_data})}: code-augmented reasoning trajectories that we constructed for complex tabular tasks, explicitly modeling the iterative think-and-execute process with code execution;
(2) \textbf{Table-oriented data}: including TableQA, tabular fact verification, table completion, code generation for table manipulation, and Text-to-SQL; and
(3) \textbf{General data}: encompassing open-domain question answering, mathematical problem solving, and general-purpose code generation.

This composition was refined through iterative benchmarking across a diverse set of evaluation suites, ensuring both strong table-specific performance and robust general reasoning capabilities. The final version data composition achieves a balanced trade-off between task specialization and broad competence, enabling the model to generalize effectively across a wide range of real-world scenarios. The distribution of different task categories is shown in Figure \ref{fig:agent_workflow}.

\section{Multi-stage Reinforcement Learning}

\subsection{Overview}
Reinforcement learning is a natural choice for improving both reasoning and generalization in tabular agents, since many realistic table tasks require iterative code execution and cannot be reliably optimized by supervised learning alone. In practice, successful RL in the tabular domain hinges on three tightly coupled components: (i) an effective optimization objective for updating the policy, (ii) a reward signal that is compatible with heterogeneous task requirements (from strict execution correctness to open-ended interpretation), and (iii) a training curriculum that prevents instability and catastrophic forgetting. In this section, we first introduce our RL optimization objective, then describe the integrated task-adaptive reward system used to compute training rewards, and finally present the multi-stage training framework that stabilizes learning.

\subsection{Agentic Data Synthesize}
\label{sec:agentic_data}
\subsubsection{Data Format}

To enable the model to learn structured, code-augmented reasoning, we require agentic training data that mimics human analytical workflows. Accordingly, each training sample is formulated as a standardized interaction trajectory: it starts with a natural language question paired with relevant tabular data, followed by an iterative think-and-execute cycle structured as follows:

\begin{enumerate}

    \item \textbf{Thinking}: The model begins by analyzing the question within a reasoning trajectory enclosed in \texttt{<think>} tags, which demonstrates problem decomposition, step-by-step planning.

    \item \textbf{Code Execution}: Executable Python code (e.g., \texttt{pd.read\_csv()}) is wrapped in \texttt{<tool\_call>} tags to invoke the Python execution environment.

    \item \textbf{Observation and Decision}: The environment returns the execution outcome (success or error) within \texttt{<tool\_response>} tags. Then the model assesses whether sufficient evidence has been gathered to answer the question. If not, it repeats this think-and-execute process, mimicking the iterative nature of real-world data analysis.

\end{enumerate}
Once confident, the model outputs a concise, self-contained response enclosed in \texttt{<answer>}. This standardized format enables reliable automatic evaluation and answer extraction.

Reinforcement learning requires precise and reliable supervision signals. Therefore, we construct reliable RL training samples via a multi-model consensus strategy. Specifically, for each query, we generate responses using multiple strong models (DeepSeek, GPT-4o, and Qwen-Max) and extract their final answers. Only samples for which at least two models produce identical final answers are retained. This cross-validation mechanism ensures high accuracy and semantic consistency in the training data, thereby enhancing the stability and effectiveness of the RL optimization process.

\begin{figure*}[!t]
    \centering
    \includegraphics[width=1.0\linewidth]{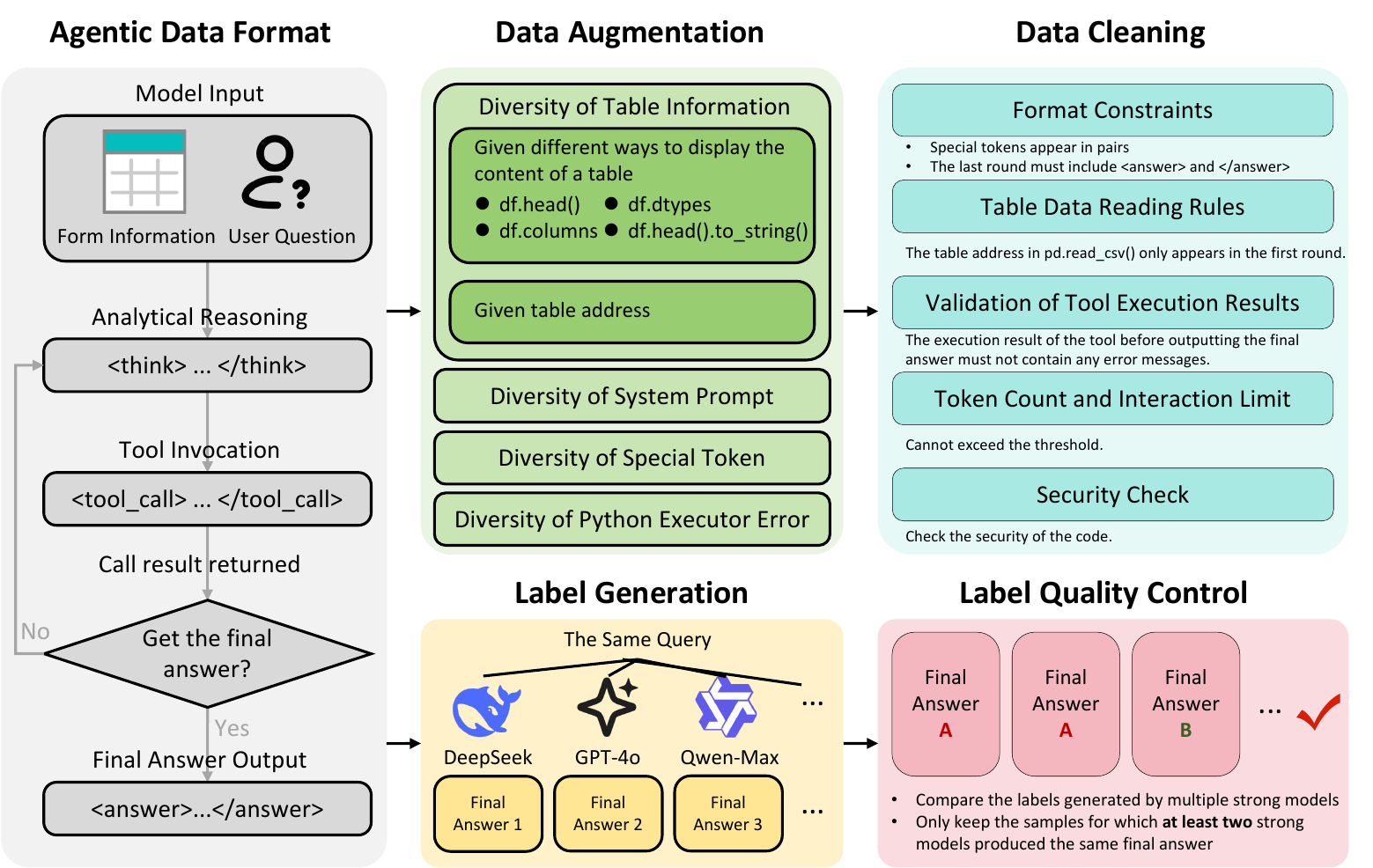}
    \caption{Agentic data synthesis pipeline for TableGPT-R1, covering structured format design, augmentation strategies, cleaning rules, and multi-model RL label validation.}
    \label{fig:agent_workflow}
\end{figure*}

\subsubsection{Data Augmentation}

To improve the model’s adaptability to diverse input scenarios and enhance the robustness of our training data, we employ multiple data augmentation strategies that specifically strengthen its tabular reasoning capabilities. A key focus of our augmentation design is \textbf{Table Input Diversity}, which directly addresses how tabular context is presented to the model. Traditionally, tables are provided in a fixed format, such as \texttt{df.head()}, \texttt{df.columns()} or schema metadata, assuming all relevant information is already available and often requiring external retrieval systems to preselect query-relevant tables, thereby limiting the model to passive reasoning over potentially incomplete snippets.

In contrast, we stochastically present tables either in conventional static formats or exclusively as \textbf{file paths}. When only a path is provided, the model must autonomously load the table, inspect its structure, and determine what information to retrieve before generating executable code. This design enables truly agentic interaction with tabular data, empowering the model to leverage its internal table comprehension capabilities to construct grounded and accurate reasoning paths. Empirically, we find that training with table-path inputs consistently achieves higher accuracy than training with static table-info representations, as the former preserves the full fidelity of the original data and allows the model to dynamically access precisely the information required for correct task execution.

Beyond Table Input Diversity, we also adopt the following augmentation techniques:
\textbf{(1) System Prompt Variation}: using diverse instruction templates to simulate varied user intents and deployment environments;
\textbf{(2) Special Token Alternation}: rotating equivalent syntactic markers such as \texttt{<function\_call>} and \texttt{<tool\_call>} to prevent overfitting to specific token patterns;
\textbf{(3) Error Message Diversity}: injecting heterogeneous Python error formats, including minimal exceptions and full stack traces, to improve robustness in diagnosing and recovering from execution failures.

\subsubsection{Reliable Label Generation}

Reinforcement learning requires precise and reliable supervision signals. Therefore, we construct reliable RL training samples via a multi-model consensus strategy. Specifically, for each query, we generate responses using multiple strong models (DeepSeek, GPT-4o, and Qwen-Max) and extract their final answers. Only samples for which at least two models produce identical final answers are retained. This cross-validation mechanism ensures high accuracy and semantic consistency in the training data, thereby enhancing the stability and effectiveness of the RL optimization process.

\subsection{Reinforcement Learning Algorithm}
We adapt Group Relative Policy Optimization (GRPO) \cite{grpo} with recent improvements introduced by DAPO \cite{dapo} and GSPO \cite{gspo}, including both sentence-level reward shaping and asymmetric (decoupled) clipping, employs the following optimization objective: 
\begin{align}
L_{\text{GRPO++}}(\theta) 
= \max \; &\mathbb{E}\left[
\frac{1}{G}
\sum_{i=0}^{G}
\min\left(
s^{(i)}(\theta) A^{(i)},
\; \text{clip}\big(s^{(i)}(\theta), 1-\varepsilon_{\text{low}}, 1+\varepsilon_{\text{high}}\big) A^{(i)}
\right)
\right] \nonumber \\
&+ C_H \, \mathbb{E}_t[H(\pi_\theta)]
- \eta \, \mathbb{E}_t\left[
\max\left(0,\; H(\pi_\theta)-H(\pi_{\theta_{\text{old}}})\right)
\right]
\end{align}
where the advantage of the $i$-th response is calculated by normalizing the group-level reward:
\begin{equation}
    A_i = \frac{r_i - \mathrm{mean}(\{r_1, r_2, \cdots, r_G\})}{\mathrm{std}(\{r_1, r_2, \cdots, r_G\})}.
\end{equation}
and the importance ratio $s^{(i)}(\theta)$ is calculated based on sentence likelihood:
\begin{equation}
s^{(i)}(\theta)
=
\exp\left(
\frac{1}{|y^{(t)}|}
\sum_{t=0}^{T}
\log \frac{\pi_\theta}{\pi_{\theta_{\text{old}}}}
\right)
\end{equation}
and to mitigate the entropy collapse commonly observed in GRPO training, we introduce two regularization terms: a high-entropy exploration bonus $C_H \, \mathbb{E}_t[H(\pi_\theta)]$ and an entropy-decay suppression term $\eta \, \mathbb{E}_t\left[ \max\left(0,\; H(\pi_\theta) - H(\pi_{\theta_{\text{old}}}) \right) \right]$, where the policy entropy is defined as
\begin{equation}
H(\pi_\theta) = -\sum_a \pi_\theta(a) \log \pi_\theta(a).
\end{equation}
It is worth noting that the high-entropy exploration bonus and the entropy-decay suppression term are not active throughout training. Based on empirical reward curves, we deactivate both terms for the first 50 steps and activate them thereafter with $C_H = \eta = 10^{-3}$, stabilizing early learning while preserving later exploration diversity.

While the objective above specifies how the policy is updated, its effectiveness critically depends on how rewards are computed for different tabular task types. We therefore next introduce the task-adaptive reward system used throughout RL training.

\subsection{Task-Adaptive Reward System}
\label{sec:Task-Adaptive Reward System}
\subsubsection{System Overview and Taxonomy}


Reinforcement learning on tabular tasks presents a distinct challenge compared to other domains: feedback signals are highly heterogeneous, spanning both deterministic objectives that admit rigorous verification (e.g., executable SQL or rule-checkable answers) and subjective objectives that require open-ended interpretation after code execution. A single reward source is therefore insufficient to reliably guide learning across the full task spectrum.
To address this, we design a Task-Adaptive Reward System whose core idea is to route terminal feedback to the most suitable verifier. For tasks with definitive correctness criteria, we employ rule-based reward functions with execution verification; for tasks without reliable ground truth, we employ a criteria-injected reward model as an LLM judge to provide calibrated quality scores. This routing policy, summarized in Table~\ref{tab:reward_taxonomy}, highlights how task properties determine the terminal evaluation pathway.
In addition to terminal feedback, tabular agents often involve multiple rounds of agentic reasoning processes. In such settings, relying solely on terminal rewards leads to brittle credit assignment. We therefore complement the task-adaptive terminal verifier with a lightweight process-level step reward that provides finer-grained signals during agentic reasoning, encouraging effective progression and improving training stability.

\begin{table}[t]
\caption{Taxonomy of Data Types and Corresponding Reward Feedback Mechanisms}
\label{tab:reward_taxonomy}
\centering
\begin{tabular}{l|c|l}
\toprule
\textbf{Task Type} & \textbf{Feedback Method} & \textbf{Remark} \\
\midrule
General & LLM-eval & Evaluated by LLM \\
Math-Logic & Rule & Rule extraction \\
Coding & Run \& Rule & Rule extraction after execution \\
Table-QA-Python & Run \& LLM-eval & Evaluated by LLM after execution \\
SQL & Run \& Rule & Rule extraction after execution \\
Table-QA-With-Label & Rule & Rule extraction \\
QA-with-Label & Rule & Rule extraction \\
\bottomrule
\end{tabular}
\end{table}


In summary, our system adaptively selects the terminal verifier (rule-based or judge-based) according to task characteristics, while additionally introducing process-level step rewards to support agentic reasoning.

\subsubsection{Reward Feedback Modules}
After defining the task-adaptive routing principle, we next describe the concrete feedback modules used to compute rewards. The key idea is to combine two complementary forms of terminal supervision---a criteria-injected reward model for open-ended tasks and a rule-based reward function for deterministically verifiable tasks---and to further introduce a lightweight process-level step reward that provides finer-grained signals for the agentic reasoning process interleaving reasoning, code execution, and environment feedback. Overall, these modules form a unified feedback toolbox that enables stable reinforcement learning across heterogeneous tabular scenarios.

\paragraph{Criteria-Injected Reward Model.}
For open-ended tabular questions that lack reliable labels, we use a criteria-injected reward model as the terminal evaluator. Rather than producing a score from implicit preference alone, the judge is conditioned on explicit evaluation criteria generated in advance by a stronger teacher model, and it scores the response by checking criterion satisfaction. As shown in Figure~\ref{fig:judge_prompt}, this design makes the reward signal more stable and explainable, and substantially reduces inconsistent scoring caused by under-specified evaluation standards. In practice, we train this judge by distilling criterion-aware evaluations from a stronger teacher model, so that the reward model learns to follow explicit standards and output consistent scores.

\begin{figure}[H]
    \centering
    \begin{center}
    \fbox{
    \begin{minipage}{0.9\linewidth}
        \small\ttfamily
        \#\# You are a reward model. Your task is to evaluate the quality of the assistant's response based on the following criteria:
        
        \textcolor{blue}{\{criteria\}}
        
        The assistant's response is as follows:
        \textcolor{blue}{\{response\}}
        
        Assign a numeric score between 0 and 10, where 0 is the worst and 10 is the best. Then provide a concise explanation for the score.
        
        \#\# Output strictly in JSON format with the following keys:
        \begin{itemize}
            \item "score": numeric value (0-10)
            \item "explanation": a brief text explaining the score
        \end{itemize}
        
        Do not include any other text outside the JSON. The explanation should be short and focused on the criteria.
        
        \#\# Example of correct output:
        
        \{\{
          "score": 8,
          "explanation": "The response is mostly accurate but misses one key detail."
        \}\}
    \end{minipage}
    }
    \end{center}
    \caption{The prompt template used for the Criteria-Injected Reward Model. The model receives pre-generated criteria and the candidate response as input context.}
    \label{fig:judge_prompt}
\end{figure}



\paragraph{Rule-Based Reward Function.}
For tasks with clear labels, we adopt deterministic terminal verification. Concretely, when the task admits exact matching or execution-based verification (e.g., SQL execution followed by rule extraction), we compute the terminal reward from the verified outcome. This rule-based method provides high-precision supervision, complementing the llm-based evaluation used for open-ended scenarios.

\paragraph{Process Step Reward.}
Beyond terminal evaluation, many tabular problems require an agentic reasoning process that interleaves reasoning, code execution, and environment feedback over multiple iterations. To provide denser guidance for this process, we introduce a lightweight rule-based step reward that scores intermediate actions according to their execution state and progress based on synthetic agentic data mentioned in section~\ref{sec:agentic_data}. Specifically, if the model selects an incorrect function, we assign a negative reward as -0.2 to discourage invalid actions. If the function selection is correct and the code execution succeeds but the episode has not yet reached the termination condition, we assign a small positive reward as +0.1 to encourage effective progression. If the function selection is correct but the code execution fails, we assign a small negative reward as -0.1 to penalize unproductive executions while still allowing recovery in subsequent steps.

\subsubsection{Reward Aggregation and Policy Regularization}
We finally aggregate the heterogeneous feedback signals into a unified reward for optimization. For an agentic reasoning trajectory $\tau$ with $T$ interaction steps, we compute the total return as the sum of per-step rewards, where policy regularization is applied at every step. Concretely, we define
\begin{equation}
R(\tau) = \sum_{t=1}^{T} \Big( R_{\text{base}}(t) + R_{\text{reg}}(t) \Big),
\end{equation}
where $R_{\text{base}}(t)$ denotes the task-adaptive reward assigned at step $t$, and $R_{\text{reg}}(t)$ is a unified regularization term applied to the model response at that step.

The step reward is state-dependent. For intermediate steps ($t<T$), $R_{\text{base}}(t)$ is given by the process step reward, which provides fine-grained signals according to execution state and progress. For the final step ($t=T$), we directly apply terminal evaluation and set $R_{\text{base}}(t) = R_{\text{terminal}}(\tau)$, which is produced by the task-adaptive verifier selected above (deterministic verification for strictly checkable tasks, or the criteria-injected judge for open-ended tasks). This formulation ensures that the final decision is always scored by the appropriate terminal verifier, while intermediate behavior is guided by lightweight process shaping.

The role of $R_{\text{reg}}(t)$ is to prevent degenerate behaviors in code-execution-centric interactions, such as unproductive self-reflection, repetitive action patterns, or overly verbose intermediate outputs that waste context. In practice, we implement $R_{\text{reg}}(t)$ as a composite penalty that captures length control, repetition suppression, and invalid-reflection mitigation (e.g., excessive ``wait''-like tokens), and we add an additional constraint to discourage opportunistic plotting. 
Additionally, to prevent reward hacking where the policy inserts unnecessary plotting code to gain credit, we introduce a simple \texttt{need\_plot} supervision signal. We annotate each training sample with a \texttt{need\_plot} indicator, and apply a strong negative penalty whenever the policy generates visualization code for samples labeled as \texttt{need\_plot=False}, thereby aligning code execution behavior with user intent and suppressing opportunistic plotting.

\begin{figure*}[!t]
    \centering
    \includegraphics[width=1.0\linewidth]{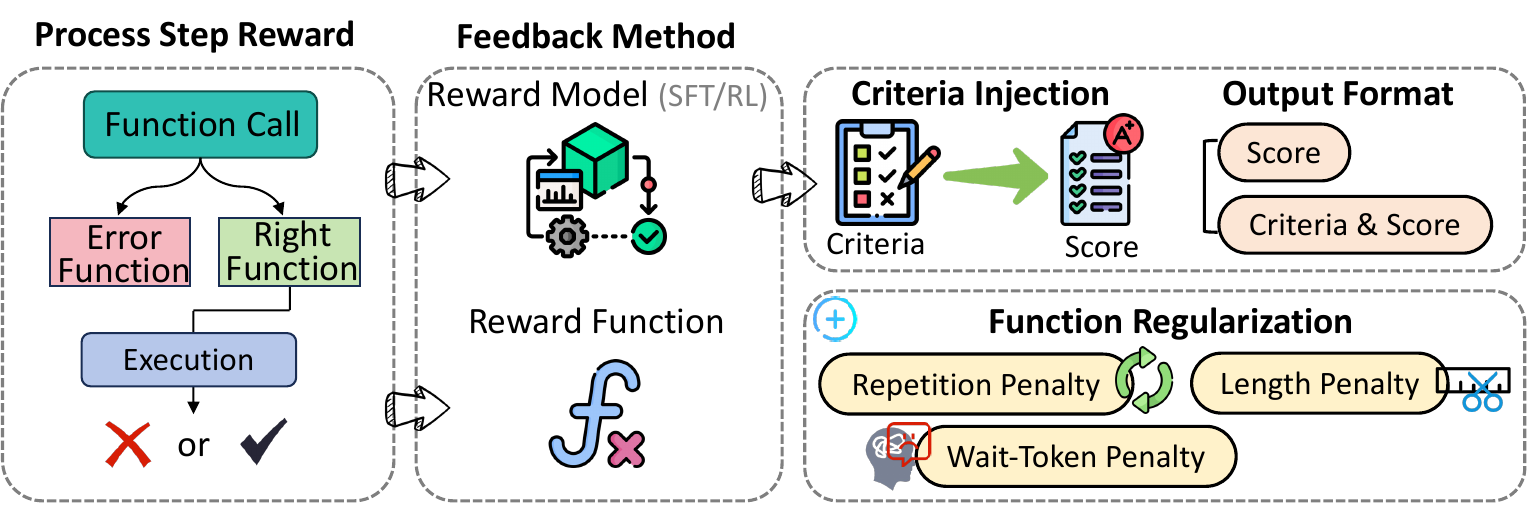}
    \caption{Illustration of task-adaptive reward system. For tasks with precise labels, we use the regularized reward function, and for open tasks, we use the reward model of criteria injection.}
    \label{fig:reward}
\end{figure*}

\subsection{Multi-stage Training Framework}
\noindent With the task-adaptive reward signal in place, we further adopt a staged training curriculum to avoid reward over-optimization and to preserve general capabilities while specializing the model for complex tabular reasoning.
\paragraph{Supervised Fine-tuning Warm-up} Due to the strict structural requirements of agentic interactions—such as explicit reasoning steps, structured code execution blocks, and formatted final answers—direct RL on the base model often produces a high rate of syntactically invalid rollouts, severely reducing training efficiency. To address this, we introduce an SFT warm-up phase before RL to reinforce consistent adherence to the target output format. Specifically, we sample approximately 3\% of the full dataset according to the final data composition ratio and fine-tune the base model (Qwen3-8B \cite{yang2025qwen3}) on this subset. This stage is not aimed at improving task performance, but rather at increasing the likelihood of generating well-formed, evaluable responses during subsequent reinforcement learning.

\paragraph{Three-stage Reinforcement Learning}

Despite this warm-up, direct RL training still exhibits unstable optimization dynamics and a tendency toward reward exploitation. In our preliminary experiments, we found two key training observations during reinforcement learning:
(1) the reward increases rapidly within the first 50 training steps, indicating quick adaptation to the reward signal;
(2) this is followed by a prolonged phase of high-frequency oscillation with minimal gains, and beyond steps 120–200, models consistently exhibit a sharp decline in general capabilities alongside various reward-hacking behaviors—such as generating corrupted tokens or skipping reasoning steps to directly output final answers. These observations suggest that without intervention, RL training tends to overfit to superficial reward cues, compromising reasoning integrity and generalization. To address this, we adopt a \textbf{three-stage} curriculum learning strategy with early stopping during reinforcement learning. Specifically:
\begin{itemize}
    \item \textbf{Stage 1}: Dominated by \textbf{general reasoning data} to establish stable reward alignment and foundational reasoning skills;
    \item \textbf{Stage 2}: Significantly increases the proportion of \textbf{Table-agentic data} to reinforce behavioral consistency in structured table tasks;
    \item \textbf{Stage 3}: Focuses exclusively on \textbf{difficult or borderline samples} (those scoring below 5 in Stage 2) to improve robustness and generalization in challenging scenarios.
\end{itemize}

We terminate each stage at step 200 and filter the data using the current policy, retaining only samples with pass@k scores of 3–7 for the next stage. This curbs reward over-optimization, bias accumulation, and overfitting. Through this multi-stage training framework, TableGPT-R1 attains enhanced structured reasoning capabilities on table-agent tasks and preserve strong general competence.

\section{Evaluation}
In this section, we present the evaluation of 8B-parameter TableGPT-R1. We first detail the experimental setup, including the constructed benchmark and the baselines used for comparison. Subsequently, we report the main results on table-related tasks and analyze the model's general capabilities.

\subsection{Experimental Setup}

\subsubsection{Benchmark Construction}

To rigorously evaluate the capabilities of TableGPT-R1, we curated a comprehensive benchmark encompassing a wide range of table-related tasks. Following prior works, we divided the benchmarks into four distinct dimensions to scrutinize the model's proficiency across diverse scenarios. The specific task classifications are structured as follows:

\begin{itemize}
    \item \textbf{Internal Benchmark}: To rigorously evaluate table question answering capabilities, we constructed a comprehensive and high-quality benchmark spanning diverse domains such as education, retail, healthcare, and finance. This dataset comprises 59 tables and 620 high-quality question-answer pairs. The construction process involved a rigorous human-in-the-loop verification pipeline: initial answers were generated by DeepSeek and subsequently scrutinized through multiple rounds of human review to eliminate ambiguity and ensure uniqueness. Uniquely, this benchmark assesses performance under two input configurations: (1) \textit{Table Info}, which provides the first $N$ rows of a table as context, and (2) \textit{Table Path}, which provides the full table's file path, thereby offering a robust evaluation of the model's adaptability.
    
    \item \textbf{Natural Language to SQL}: To evaluate the model's ability to translate natural language queries into executable SQL statements, we selected the most widely recognized benchmarks, BIRD~\cite{li2024bird} and Spider~\cite{yu2018spider}. These datasets serve as the standard for assessing semantic parsing and generating correct SQL code.
    
    \item \textbf{Holistic Table Evaluation}: We employed the TableBench~\cite{wu2024tablebench} to test general analytical skills. This dataset spans 18 industry domains and encompasses four key analysis tasks: fact verification, numerical reasoning, data analysis, and code-based chart visualization, providing a holistic assessment of tabular comprehension. Besides, to evaluate the model's reasoning performance on irregular tables, we introduced RealHitBench~\cite{wu2025realhitbenchcomprehensiverealistichierarchical}, the latest dataset in this domain. As a challenging benchmark featuring hierarchical and complex tables, it presents data in multiple modalities and formats. Furthermore, it includes a variety of question types designed to rigorously test the model's reasoning capabilities on intricate, real-world data structures.
    
    \item \textbf{Agent-based Data Analysis}: To assess the model's capability in tabular data analysis, we utilized InfiAgent-DABench, a benchmark tailored for evaluating LLM-based agents. This framework rigorously tests the model's capacity to resolve complex, end-to-end data analytics challenges by dynamically interacting with an execution environment.
\end{itemize}

\subsubsection{Baselines}
For comparative analysis, we benchmarked our \textbf{TableGPT-R1} against a wide spectrum of Large Language Models (LLMs), categorized into three distinct groups: (1) \textbf{State-of-the-art Open-source LLMs}, including general-purpose models such as the \textbf{Qwen3 Series}~\cite{yang2025qwen3}, \textbf{Llama-3.1-8B-Instruct}~\cite{dubey2024llama}, and \textbf{QwQ-32B}~\cite{qwq32b}; (2) \textbf{Leading Proprietary LLMs}, representing top-tier closed-source capabilities, such as \textbf{GPT-4o}~\cite{hurst2024gpt}, \textbf{DeepSeek-Chat}~\cite{deepseekai2024deepseekv3technicalreport}, and \textbf{Qwen-Plus}~\cite{yang2025qwen3}; and (3) \textbf{Specialized Table LLMs}, specifically \textbf{TableLLM}~\cite{wu2024tablebench}, \textbf{
Table-R1-Zero-7B}~\cite{yang-etal-2025-table} and \textbf{TableGPT2-7B}, which are explicitly fine-tuned or optimized for tabular intelligence.

\subsection{Main Results on Table-Specific Tasks}
\begin{table*}[!t]
    \centering
    \caption{Performance comparison grouped by model scale. \textbf{Left Group}: Models with comparable parameters to TableGPT-R1-8B. \textbf{Right Group}: Significantly larger models and proprietary closed-source models. \textbf{Bold} indicates the best result within each group. \textbf{Gray background} highlights TableGPT-R1-8B.
    \textbf{Abbreviations:} \textbf{Q3}: Qwen3; \textbf{QwQ}: QwQ-32B; \textbf{DS-V3}: DeepSeek-V3; \textbf{Q-Plus}: Qwen-Plus; \textbf{T-LLM}: TableLLM; \textbf{Llama}: Llama-3.1-8B; \textbf{T-R1-Z}: Table-R1-Zero-7B; \textbf{TGPT2}: TableGPT2-7B; \textbf{TGPT-R1}: TableGPT-R1-8B; \textbf{FC}: Fact Checking; \textbf{NR}: Numerical Reasoning; \textbf{SC}: Structure Comprehending; \textbf{DA}: Data Analysis; \textbf{CG}: Chart Generation.}
    \label{tab:main_results}
    
    \definecolor{highlightgray}{gray}{0.95}
    
    \renewcommand{\arraystretch}{1.25} 
    \setlength{\tabcolsep}{2pt} 
    
    \resizebox{\textwidth}{!}{%
    \begin{tabular}{ll c | ccccc >{\columncolor{highlightgray}}c | ccccccc | cc}
        \toprule
        \multirow{2}{*}{\textbf{Benchmark}} & \multirow{2}{*}{\textbf{Task}} & \multirow{2}{*}{\textbf{Met.}} & \multicolumn{6}{c|}{\textbf{Comparable Parameter Models }} & \multicolumn{7}{c|}{\textbf{Larger Parameter \& Proprietary Models}} & \multicolumn{2}{c}{\textbf{Improv.}} \\
        \cmidrule(lr){4-9} \cmidrule(lr){10-16} \cmidrule(lr){17-18}
        & & &  Q3-8B &  T-LLM &  Llama & T-R1-Z & TGPT2 &  \textbf{TGPT-R1} &  Q3-14B &  Q3-32B &  Q3-30B &  QwQ &  GPT-4o &  DS-V3 &  Q-Plus &  vs.Q3-8B &  vs.TGPT2 \\
        \midrule
        
        \multicolumn{18}{c}{\textit{Internal Bench}} \\
        \midrule
        \multicolumn{2}{l}{Table Info} & Acc & 69.20 & 0.97 & 37.26 & 15.97 & - & \textbf{80.00} & 66.10 & 72.58 & 51.10 & 69.68 & 67.26 & 66.00 & \textbf{76.90} & 10.80 & - \\        
        \multicolumn{2}{l}{Table Path} & Acc & 73.90 & 0.65 & 31.77 & 9.19 & - & \textbf{82.70} & 74.70 & 78.55 & 60.50 & 75.00 & - & 72.90 & \textbf{81.50} & 8.80 & - \\
        \midrule

        \multicolumn{18}{c}{\textit{NL2SQL}} \\
        \midrule
        \multicolumn{2}{l}{Spider} & EX & 86.07 & 65.30 & 73.59 & 82.63 & 74.38 & \textbf{86.73} & 87.61 & 87.80 & 61.71 & 85.33 & 87.98 & 88.54 & \textbf{89.19} & 0.66 & 12.35 \\
        \multicolumn{2}{l}{BIRD} & EX & 61.67 & 30.64 & 40.03 & 50.98 & 49.28 & \textbf{63.17} & 61.80 & 63.04 & 53.91 & 54.30 & 65.25 & 65.65 & \textbf{68.32} & 1.50 & 13.89 \\
        \midrule
        
        \multicolumn{18}{c}{\textit{Holistic Table Evaluation}}\\
        \midrule
        \multicolumn{2}{l}{TableBench} & & & & & & & & & & & & & & & & \\
        & DP & Rge & 42.10 & 3.63 & 18.04 & 39.4 & 42.10 & \textbf{48.35} & 47.41 & \textbf{52.18} & 48.61 & 49.33 & 40.91 & 36.56 & 31.01 & 6.25 & 6.25 \\
        & PoT & Rge & 28.01 & 0.00 & 6.73 & 7.54 & \textbf{39.80} & 35.12 & 36.61 & 37.78 & 27.72 & 40.03 & \textbf{51.96} & 33.05 & 41.79 & 7.11 & -4.68 \\
        & SCoT & Rge & 41.86 & 1.99 & 21.94 & 28.89 & 40.70 & \textbf{49.53} & 47.36 & 47.47 & 45.68 & 44.84 & 41.43 & \textbf{50.11} & 44.06 & 7.67 & 8.83 \\
        & TCoT & Rge & 41.71 & 3.18 & 15.26 & 39.52 & 46.19 & \textbf{48.28} & 46.07 & 51.74 & 47.63 & 48.83 & 45.71 & \textbf{54.28} & 52.07 & 6.57 & 2.09 \\
        
        \multicolumn{2}{l}{RealHitBench} & & & & & & & & & & & & & & & & \\
        & FC & EM & 58.83 & 33.44 & 30.32 & 0.00 & 43.06 & \textbf{63.85} & 62.36 & 65.00 & 60.23 & \textbf{66.31} & 55.22 & 65.08 & 56.53 & 5.01 & 20.79 \\
        & NR & EM & 39.43 & 13.36 & 14.53 & 0.00 & 24.90 & \textbf{49.03} & 43.70 & 47.34 & 46.95 & \textbf{55.38} & 38.91 & 52.53 & 31.25 & 9.60 & 24.13 \\
        & SC & EM & 64.12 & 53.28 & 35.90 & 28.50 & 34.86 & \textbf{64.12} & 73.02 & 71.76 & 69.47 & \textbf{76.08} & 61.83 & 71.25 & 62.85 & 0.00 & 29.26 \\
        & DA. & GPT & 53.28 & 47.86 & 60.12 & 36.24 & 53.16 & \textbf{66.53} & 63.03 & \textbf{66.67} & 53.27 & 64.99 & 55.54 & 66.29 & 62.04 & 13.25 & 13.37 \\
        & CG & ECR & 24.67 & 22.73 & 13.64 & 16.00 & 44.16 & \textbf{55.84} & 23.38 & 25.00 & 20.78 & 20.13 & 34.42 & 18.18 & \textbf{48.05} & 31.17 & 11.68 \\
        \midrule
        
        \multicolumn{18}{c}{\textit{Agent-based Data Analysis}}\\
        \midrule
        \multicolumn{2}{l}{InfiAgent-DA} & Acc & 56.81 & 11.67 & 55.08 & 70.82 & 73.15 & \textbf{80.54} & 59.92 & 54.86 & 41.63 & 37.74 & \textbf{87.10} & 77.43 & 67.32 & 23.73 & 7.39 \\        
        \bottomrule
        
    \end{tabular}%
    }
\end{table*}
Table~\ref{tab:main_results} presents the comprehensive comparative analysis of TableGPT-R1 against the state-of-the-art LLMs discussed above. Notably, even without extensive training on specific benchmark training sets, our 8B-parameter TableGPT-R1 achieves significantly superior results compared to nearly all other Table-oriented Models. Empirically, on select benchmarks, it delivers performance comparable to, or even exceeding, that of proprietary models like GPT-4o.

Overall, TableGPT-R1 demonstrates substantial advancements over its predecessor, TableGPT2-7B, particularly in table comprehension and reasoning capabilities. Detailed comparisons are as follows:

On the TableBench benchmark, TableGPT-R1 demonstrates strong performance. It achieves an average gain of $6.9\%$ over the Qwen3-8B across four core sub-tasks. Compared to the TableGPT2-7B, it records an average improvement of $3.12\%$, validating its enhanced reasoning capability despite a trade-off in the PoT task.

In the Natural Language to SQL domain, TableGPT-R1 exhibits superior generalization capabilities. While showing consistent improvements over Qwen3-8B on Spider 1.0 ($+0.66\%$) and BIRD ($+1.5\%$), it represents a significant leap compared to TableGPT2-7B, registering dramatic performance increases of $12.35\%$ and $13.89\%$, respectively.

In the highly challenging RealHitBench test,TableGPT-R1 achieved outstanding results, particularly surpassing the top closed-source baseline model GPT-4o. This highlights its powerful capabilities in hierarchical table reasoning. Quantitative analysis shows that TableGPT-R1 matches or outperforms Qwen3-8B across subtasks, achieving an average improvement of $11.81\%$, with a remarkable peak gain of $31.17\%$ in the Chart Generation task. Furthermore, compared to TableGPT2-7B, the model represents a significant advancement, registering an average improvement of $19.85\%$ across all subtasks.

Evaluation on our Internal Benchmark further attests to the model's robustness. TableGPT-R1 surpasses Qwen3-8B by substantial margins: $10.8\%$ on the Table Info and $8.8\%$ on the Table Path.

\begin{table*}[!t]
    \centering
    \caption{Evaluation of general capabilities across coding, mathematics, and general knowledge benchmarks. \textbf{Model Abbreviations:} \textbf{Q3}: Qwen3 Series; \textbf{QwQ}: QwQ-32B; \textbf{T-LLM}: TableLLM; \textbf{TGPT2}: TableGPT2-7B; \textbf{TGPT-R1}: TableGPT-R1-8B (Ours). \textbf{vs. Q3-8B}: Improvement over Qwen3-8B; \textbf{vs. TGPT2}: Improvement over TableGPT2-7B. Best results are highlighted in bold.}
    \label{tab:general_results}
    \definecolor{highlightgray}{gray}{0.95}
    \renewcommand{\arraystretch}{1.2}
    \setlength{\tabcolsep}{3pt}
    
    \resizebox{\textwidth}{!}{%
    \begin{tabular}{l ccccc | cc >{\columncolor{highlightgray}}c | cc}
        \toprule
        \multirow{2}{*}{\textbf{Benchmark}} & \multicolumn{5}{c|}{\textbf{General Open-Source Baselines}} & \multicolumn{2}{c|}{\textbf{Specialized}} & \textbf{Ours} & \multicolumn{2}{c}{\textbf{Improvement}} \\
        \cmidrule(lr){2-6} \cmidrule(lr){7-8} \cmidrule(lr){9-9} \cmidrule(lr){10-11}
        & \small Q3-8B & \small Q3-14B & \small Q3-32B & \small Q3-30B & \small QwQ & \small T-LLM &  \small TGPT2 & \small \textbf{TGPT-R1} & \small vs. Q3-8B & \small vs. TGPT2 \\
        \midrule
        
        \multicolumn{11}{c}{\textit{Coding Capabilities}} \\
        \midrule
        HumanEval & 93.90 & 95.73 & 95.73 & 95.12 & \textbf{96.34} & 18.29 & 80.00 & 95.73 & 1.83 & 15.73 \\
        MBPP & 84.80 & 80.60 & 79.20 & 87.60 & \textbf{87.80} & 15.60 & 64.00 & 83.20 & -1.60 & 19.20 \\
        \midrule
        
        \multicolumn{11}{c}{\textit{Mathematical Reasoning}} \\
        \midrule
        GSM8K & 93.70 & 92.90 & 95.20 & 95.10 & 88.80 & 68.70 & 83.10 & \textbf{95.60} & 1.90 & 12.50 \\
        MATH & 91.00 & 90.20 & 91.90 & 78.80 & 84.20 & 29.20 & 44.50 & \textbf{93.30} & 2.30 & 48.80 \\
        AIME & 40.00 & 56.67 & \textbf{60.00} & \textbf{60.00} & 50.00 & 0.00 & 3.33 & 50.00 & 10.00 & 46.67 \\
        \midrule
        
        \multicolumn{11}{c}{\textit{General Knowledge}} \\
        \midrule
        CMMLU-5 & 79.66 & 85.33 & \textbf{86.97} & 85.90 & 63.39 & 49.50 & 79.45 & 76.84 & -2.82 & -2.61 \\
        GPQA-Diamond & 60.10 & 59.60 & \textbf{65.66} & 58.08 & 57.58 & 6.57 & 16.67 & 55.56 & -4.54 & 38.89 \\

        \bottomrule
    \end{tabular}%
    }
\end{table*}

\subsection{Case Study}
\begin{figure*}[!t]
    \centering
    \includegraphics[width=1.0\linewidth]{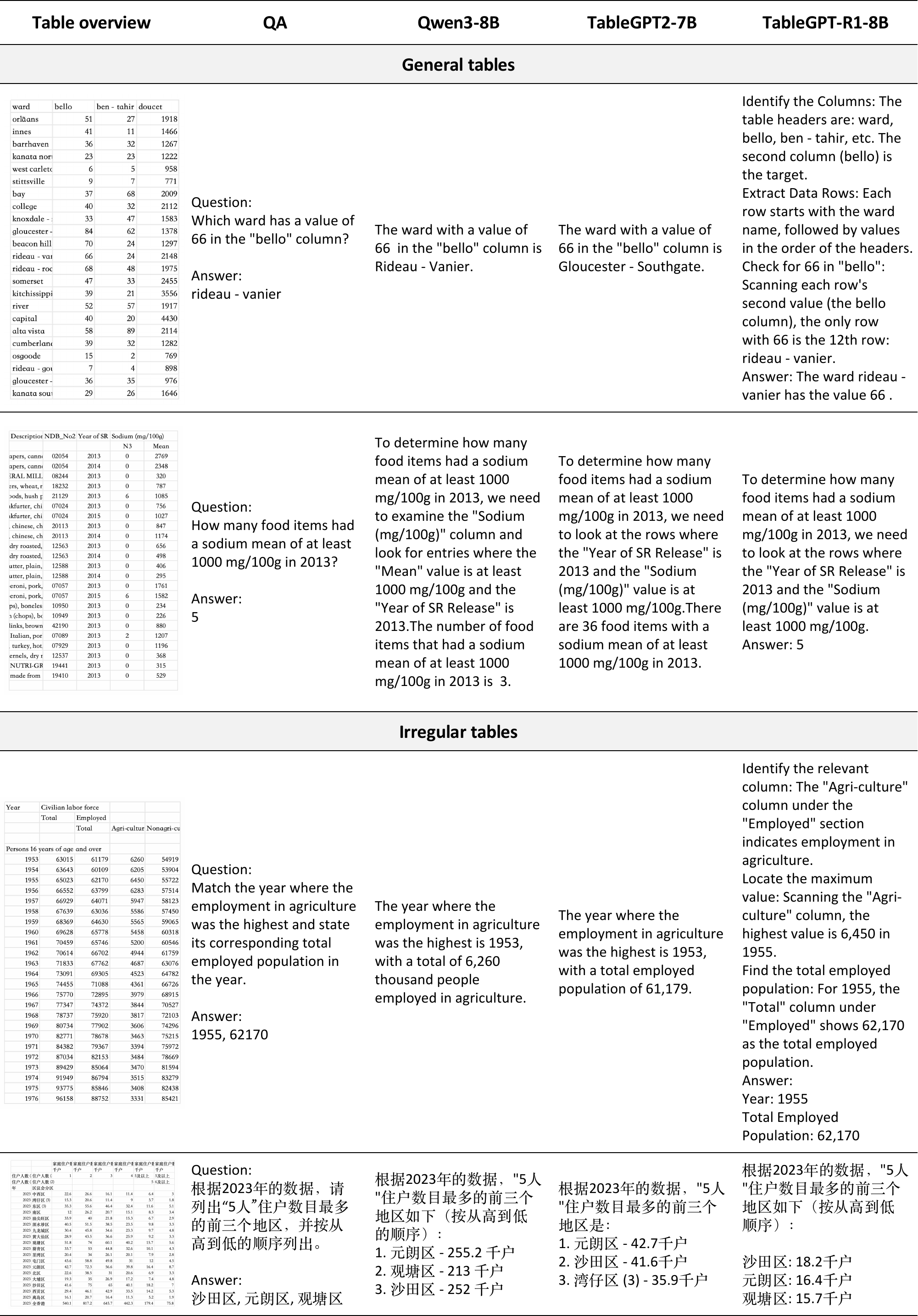}
    \caption{Qualitative cases of TableGPT-R1.}
    \label{fig:case_study}
\end{figure*}

To provide a tangible demonstration of TableGPT-R1's reasoning capabilities on tabular data, we present qualitative examples derived from our evaluation dataset in Figure~\ref{fig:case_study}. Each case includes the source table, the specific query, and the response generated by model. By contrasting these outputs with those from baseline models under identical conditions, we empirically highlight the superior accuracy and logical coherence of TableGPT-R1 in handling complex tabular scenarios.

\subsection{General Capabilities}

A critical objective in specializing Large Language Models for tabular tasks is to enhance domain-specific expertise without compromising general intelligence. To verify that TableGPT-R1 avoids catastrophic forgetting, we evaluated it on a suite of widely recognized benchmarks, including MBPP\cite{austin2021program}, HumanEval\cite{chen2021evaluating}, GSM8K\cite{cobbe2021gsm8k}, MATH\cite{hendrycksmath2021}, CMMLU-5\cite{li2023cmmlu}, GPQA-Diamond\cite{rein2023gpqagraduatelevelgoogleproofqa}, and AIME\cite{patel2024aimeaioptimizationmultiple}. These datasets comprehensively assess general coding proficiency, mathematical reasoning, and language understanding.

The results, presented in Table~\ref{tab:general_results}, demonstrate that TableGPT-R1 maintains excellent general capabilities. Despite extensive optimization for table-related tasks, the model exhibits high stability in general coding and mathematical reasoning. 

Quantitatively, TableGPT-R1 shows an average performance gain of $1.01\%$ across the seven metrics compared to the base model, Qwen3-8B, while achieving a substantial average improvement of $25.6\%$ over its predecessor, TableGPT2-7B. Specifically, TableGPT-R1 demonstrates improvements over Qwen3-8B on HumanEval, GSM8K, and MATH, and achieves a notable increase on AIME ($+10.0\%$). Minor performance regressions were observed on MBPP, CMMLU-5, and GPQA-Diamond.

\section{Conclusion}
In this work, we presented \textbf{TableGPT-R1}, a specialized tabular model that effectively bridges the gap between general-purpose reasoning and domain-specific data analysis through a systematic reinforcement learning framework. By addressing the critical challenges of data scarcity, feedback heterogeneity, and optimization instability, our approach establishes a robust paradigm for vertical domain adaptation. The proposed \textbf{Task-Adaptive Reward System}—which synergizes rule-based verification for rigid tasks with criteria-injected reward modeling for open-ended reasoning—alongside a \textbf{Multi-Stage Training Strategy}, has proven highly effective in stabilizing training and preventing catastrophic forgetting.
Experimental results across various benchmarks demonstrate that TableGPT-R1 achieves state-of-the-art performance in table comprehension, code execution, and complex reasoning, significantly outperforming comparable open-source models while retaining robust general intelligence. These findings validate that a rigorously designed RL pipeline can enhance the model to expert-level proficiency in specialized domains while maintain general performance. Future work will focus on extending these capabilities to multi-modal table understanding and scaling to ultra-large database interactions, further pushing the boundaries of autonomous data analysis.

\section*{Acknowledgements}
This work is supported by the National Regional Innovationand Development Joint Fund (No. U24A20254). This work was partially supported by ZJU Kunpeng$\&$Ascend Center of Excellence.
\clearpage

\bibliography{neurips_2025}
\bibliographystyle{plain}


\end{document}